\documentclass[runningheads]{llncs}
\usepackage{graphicx}
\usepackage{xcolor}
\usepackage{amssymb}
\usepackage{amsmath}
\usepackage{multirow}
\usepackage[linesnumbered,ruled]{algorithm2e}
\usepackage{tabularray}
\usepackage{multirow}
\usepackage{lscape}
\usepackage{booktabs}
\begin{document}

\title{Security Assessment of Hierarchical Federated Deep Learning}

\author{Anonymous submission}
%
%
\author{Duaa S. Alqattan\inst{1,2}
\and
Rui Sun\inst{1} \and
Huizhi Liang\inst{1} \and
Guiseppe Nicosia\inst{4} \and
Vaclav Snasel\inst{3}\and
Rajiv Ranjan\inst{1} \and
Varun Ojha\inst{1}
}

\authorrunning{D. S. Alqattan et al.}
%
\institute{Newcastle University, Newcastle, UK \and
Alahsa Technical College, Technical and Vocational Training Corporation, Alahsa, Saudi Arabia \and
Technical University of Ostrava, Ostrava, Czech Republic \and
University of Catania, Catania, Italy
}
\maketitle              
\begin{abstract}
Hierarchical federated learning (HFL) is a promising distributed deep learning model training paradigm, but it has crucial security concerns arising from adversarial attacks. This research \textit{investigates and assesses the security of HFL} using a novel methodology by focusing on its resilience against inference-time and training-time adversarial attacks. Through a series of extensive experiments across diverse datasets and attack scenarios, we uncover that HFL demonstrates robustness against untargeted training-time attacks due to its hierarchical structure. However, targeted attacks, particularly backdoor attacks, exploit this architecture, especially when malicious clients are positioned in the overlapping coverage areas of edge servers. Consequently, HFL shows a dual nature in its resilience, showcasing its capability to recover from attacks thanks to its hierarchical aggregation that strengthens its suitability for adversarial training, thereby reinforcing its resistance against inference-time attacks. These insights underscore the necessity for balanced security strategies in HFL systems, leveraging their inherent strengths while effectively mitigating vulnerabilities.

\end{abstract}

\section{Introduction}
\begin{figure}[t!]
\centering
\includegraphics[width=12cm]{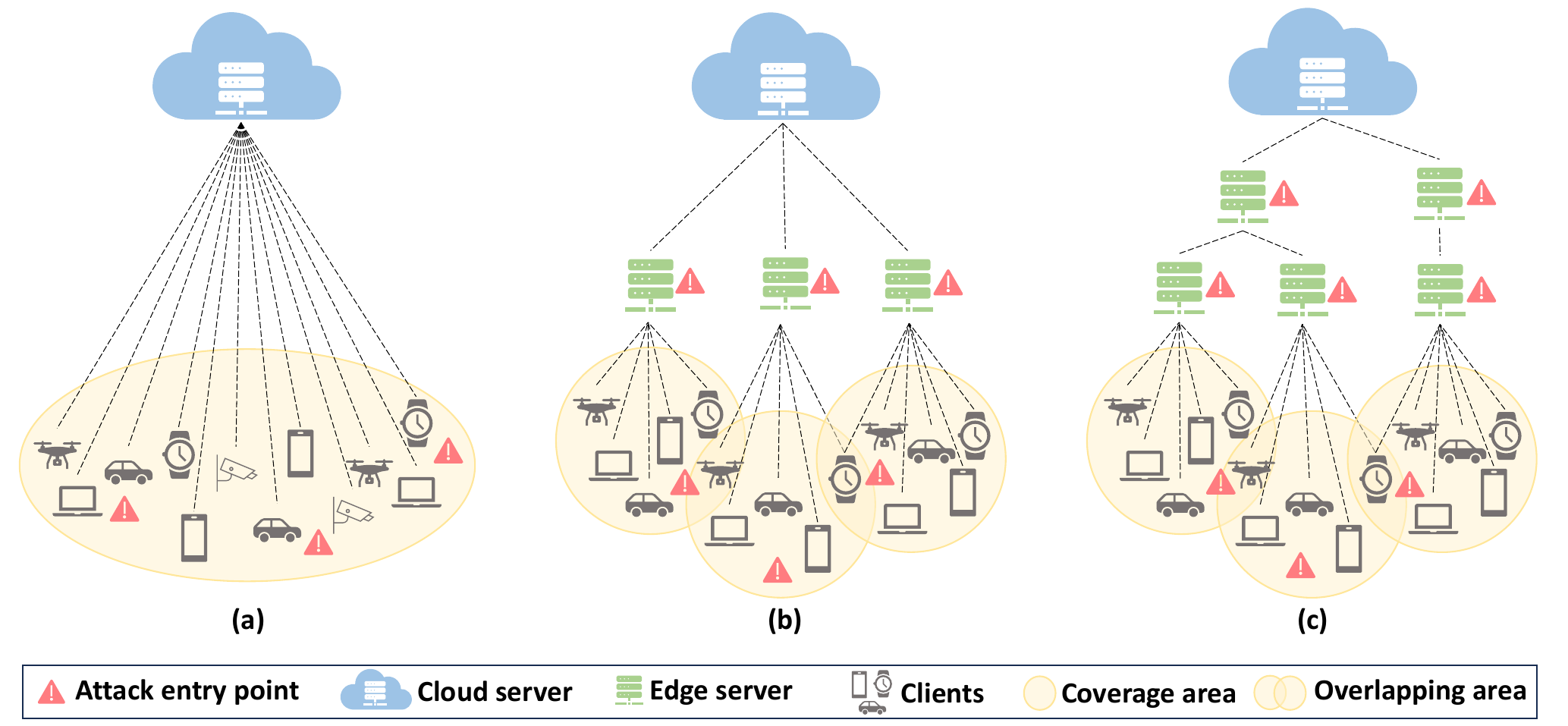}
\caption{FL network architectures: (a) 2-level FL; (b) 3-level HFL; (c) 4-level HFL}
\label{fig:HFLoverview}
\end{figure}

Federated Learning (FL) offers a promising solution to the challenges of Centralized Machine Learning (CML), including data storage, computation, and privacy. FL facilitates collaborative training of a global model across numerous clients while preserving data decentralization. This approach has been successful in various applications like smart cities. Traditionally, FL employed a two-level node design, where chosen clients submit updates to a central server, situated either at the \textit{edge} or in the \textit{cloud}, for aggregation, as shown in Fig~\ref{fig:HFLoverview}(a). The aggregation at the edge improves latency and network efficiency but restricts server capacity, affecting training. The aggregation in the cloud boosts computational power and scalability but may delay updates for distant devices, stressing networks. In recent years, hierarchical federated learning (HFL), a variant of FL, has gained attention. HFL addresses FL challenges by employing multiple aggregator servers at edge and cloud levels, hierarchically interconnected, capitalizing on cloud coverage, and reducing the edge server latency~\cite{yan2023hierarchical}.

In a use case scenario where HFL is deployed for smart city applications such as image classification, various clients, including smart cars, smart watches, drones, and mobile phones, are scattered across the smart city ~\cite{rana2022hierarchical}. A significant number of edge servers are typically deployed in close proximity to these clients, forming a distributed architecture network connected to a central cloud server. Clients establish connections with edge servers within their coverage areas, with overlapping coverage enabling connections to multiple edge servers~\cite{han2021fedmes,qu2022convergence}. These edge servers forward client updates to regional edge servers, ultimately reaching the cloud server for aggregation to build a global model. Fig.~\ref{fig:HFLoverview} shows a comparison of 2-level FL (Fig.~\ref{fig:HFLoverview}(a)) and HFL architectures that can be employed as 3-level~\cite{liu2020client} (Fig.~\ref{fig:HFLoverview} (b)) and 4 level node design~\cite{zhou2023toward} (Fig.~\ref{fig:HFLoverview}(c)). 


Despite the advantages of HFL over FL, HFL remains susceptible to adversarial attacks that compromise data integrity by manipulating local datasets or model updates to undermine the global model's performance~\cite{ooi2023measurement}. In HFL architecture, the increased number of nodes, including clients and edge servers, expands the attack surface, providing more potential entry points for attacks. This amplifies the risk of compromises by malicious edge servers or clients, surpassing the attack surface of FL. Fig.~\ref{fig:HFLoverview} provides an overview of the attack surface (see red triangle) in conventional FL compared to  HFL. However, the augmentation of nodes also presents opportunities for bolstering defense mechanisms against attacks. This prompts an exploration of the following question: \textit{How does the HFL architecture impact the robustness of HFL against attacks}?

While previous studies have evaluated 3-level HFL model convergence~\cite{liu2020client,liu2022hierarchical} and proposed resilient aggregation methods for 4-level HFL models~\cite{zhou2023toward}, based on our best knowledge, there is little to no work on a systematic assessment of HFL security available in the literature that we aim to do in this paper. We examined HFL's resilience to adversarial attacks in detail. With the growing use of HFL in smart city applications~\cite{ooi2023measurement}, it is crucial to evaluate their resilience and understand their architectural nuances to suggest areas for improvement.

This paper explores how the HFL architecture withstands adversarial data injected during inference. Our findings highlight the challenges inference-time attacks pose to model accuracy. Yet, defense strategies like adversarial training offer promising solutions. We delve into Data Poisoning Attacks (DPA) and Model Poisoning Attacks (MPA) at the client and server sides during training, alongside potential defense mechanisms within the HFL framework. We identify vulnerabilities to targeted DPA (backdoor attack), notably in the 4-level HFL model, where hierarchical structure affects malicious client selection probabilities. Implementing the neural cleanser method~\cite{nicolae2018adversarial} proves effective against targeted backdoor attacks, emphasizing tailored defense strategies' importance. Conversely, HFL models show resilience against untargeted DPA and MPA due to multi-level aggregation, mitigating outlier impact and enabling recovery from attacks.

In summary, our contributions are as follows:
\begin{enumerate}
    \item We present a novel methodology for assessing the security of HFL that offers insights into the resilience of HFL against inference time attacks, enhancing our understanding of HFL's robustness. 
    
    \item Through comparative analyses, we pinpoint vulnerabilities in HFL under various training-time attacks and investigate how the HFL architecture influences model resilience against attacks, deepening our understanding of FL design and security.     
    
    \item Our assessment of adversarial hierarchical federated training via extensive experiments on different datasets and HFL architectures sheds light on effective defense mechanisms for future HFL framework development, emphasizing HFL's resilience and its capacity to recover from attacks.
\end{enumerate}


\section{Related Work}
\label{sec:Related_work}
In recent years, significant attention has been devoted to studying the impact of attacks on FL. Abyane et al.~\cite{abyane2023towards} conducted an empirical investigation to comprehensively understand the quality and challenges associated with state-of-the-art FL algorithms in the presence of attacks and faults. Shejwalkar et al.~\cite{shejwalkar2022back} systematically categorized various threat models, types of data poisoning, and adversary characteristics in FL, assessing the effectiveness of these threat models against basic defense measures. Bhagoji et al.~\cite{bhagoji2019analyzing} explored the emergence of model poisoning, a novel risk in FL, distinct from conventional data poisoning. 

In contrast to conventional 2-level FL, adopting HFL introduces many novel research concerns due to its inherently intricate multi-level design~\cite{yan2023hierarchical}. A few studies have focused on examining convergence in HFL~\cite{liu2020client,liu2022hierarchical}. Some studies offer solutions to some of the issues related to HFL security. Zhou et al.~\cite{zhou2023toward} introduced a robust model aggregation technique aimed at ensuring the resilience of 4-level HFL against poisoning attacks, particularly in the context of the Internet of Vehicles (IoV). Al-Maslamani et al.~\cite{al2023reputation} tackled the issue of selecting unreliable clients within the 3-level HFL framework to optimize overall HFL security. To the best of our knowledge, scholarly works assessing the security aspects of HFL are relatively scarce. In comparison to these studies, our research focuses on conducting a systematic assessment of the security of HFL. 

\section{Security Assessment of Hierarchical Federated Learning}
\label{sec:method}
\subsection{Hierarchical Federated Learning (HFL) Model} \label{HFL_Model}
We conceptualize the HFL system as a multi-parent hierarchical tree (as shown in Fig.~\ref{fig:HFLoverview}), denoted as $T=(V,E)$, consisting of $|L|$ levels. Nodes in the system, categorized as clients ($N$) and servers ($S$), are represented in the set $V$, while the collection of undirected communication channels between nodes is represented in the set $E$. The cloud server node, $s_0$, serves as the root of the tree at level $0$, with client nodes, $n$, positioned at the leaves of the tree at level $L-1$. Intermediate edge servers, $s_\ell$, act as intermediary nodes between cloud servers and clients at level $\ell$ ($\ell \in \{1,\ldots, L-2\}$). Clients may train their local models using local data and transmit their model parameters to regional edge servers $s_{L-2}$ for aggregation. The aggregation process in an HFL system involves several critical steps shown in figure \ref{fig:AttackModel}. (\textit{Step 1}) The cloud server $s_0$ sends the initial model to clients $n$ through edge servers $s_\ell$. (\textit{Step 2}) Regional edge servers $s_{L-2}$ select a set of client participants $C_t$ at aggregation round $t$ from their coverage areas $A(s_{L-2})$ for model updates. (\textit{Step 3}) Clients $C_t$ download the latest model from regional edge servers $s_{L-2}$ and train their local models. (\textit{Step 4}) Updated parameters are sent back to regional edge servers $s_{L-2}$ for aggregation. (\textit{Step 5}) Parent servers $s_\ell$ at level $\ell$ aggregate updated model parameters from child nodes $s_\ell+1$ within their coverage areas $A(s_\ell)$ for $T_\ell$ Number of aggregation rounds. (\textit{Step 6}) After $T_0$ global aggregation rounds implemented by cloud server $s_0$, a global model is constructed and transmitted to clients for deployment through edge servers $s_\ell$.








We employ the averaging aggregation method proposed by McMahan et al.~\cite{mcmahan2017communication}, allowing flexibility in deploying HFL models with varying levels ($L$). 

\subsection{Adversarial Attacks on HFL Model} \label{Attack_model}
We consider the attacks on HFL models targeting data integrity during both the \textit{training} and \textit{inference} time. These attacks can be client-side or server-side, with client-side attacks encompassing data poisoning and model poisoning tactics.

\begin{figure}[!t]
\centering
\includegraphics[width=\textwidth]{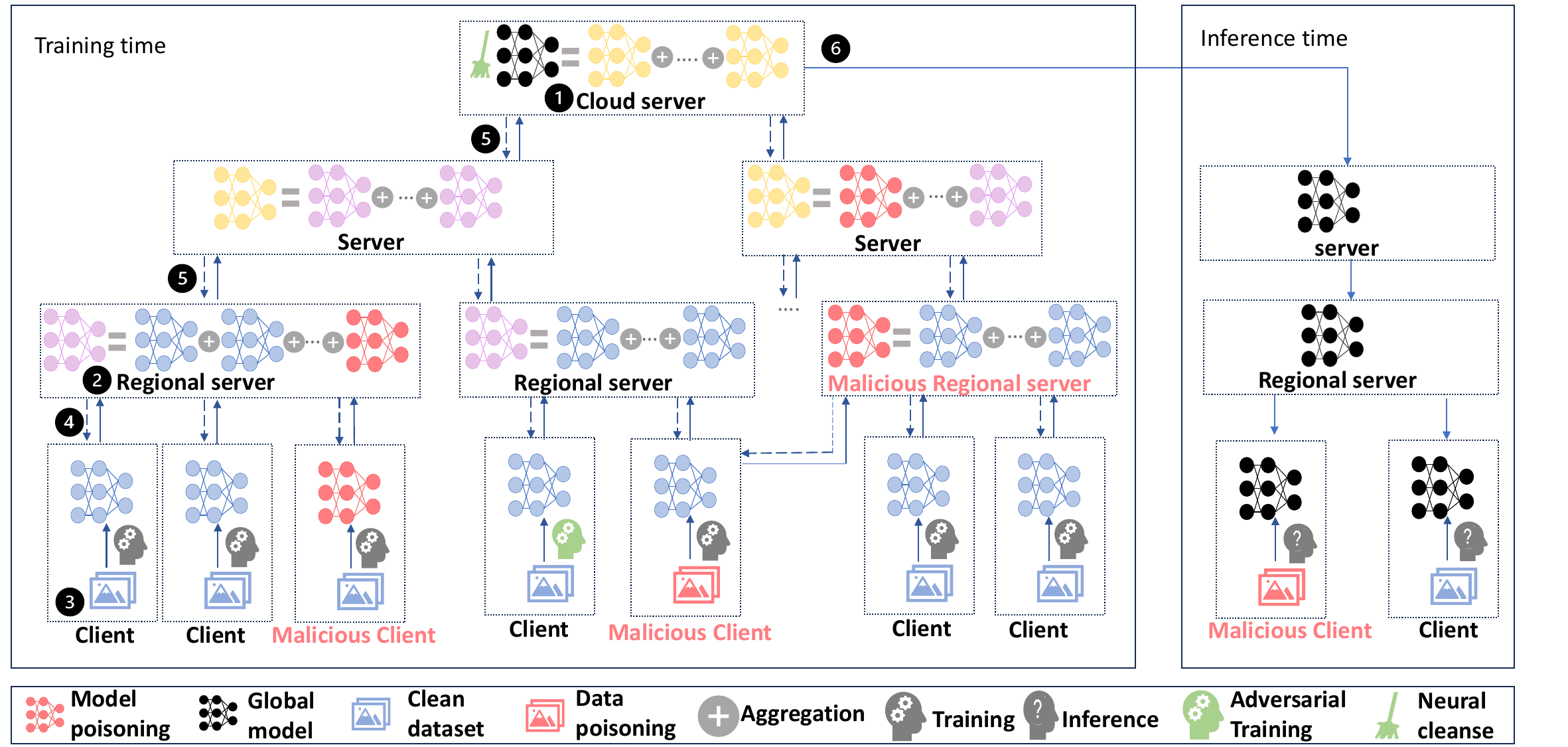}
\caption{HFL and Attack Model}
\label{fig:AttackModel}
\end{figure}

\paragraph{Inference-time Attacks (ITAs).} ITAs aim to carefully perturb the input data at inference time to have them misclassified by the global model. Adversarial data is created through two types of ITAs: white-box attacks and black-box attacks, determined by the attacker's access level to the target global model. White-box attacks require full access to the target model, including its architecture, parameters, and gradients. Black-box attacks, on the other hand, do not rely on or require access to the internal details of the target global model. In this work, we have applied white-box attacks, including Adversarial Patch(AP), Fast Gradient Method(FGM), Projected Gradient Descent(PGD), and Saliency Map Method(JSMA). We also applied black-box attacks, including Square Attack(SA) and Spatial Transformations Attack(ST)~\cite{nicolae2018adversarial}.

\paragraph{Training-time Attacks (TTAs).} TTAs aim to inject adversarial data during training time to influence model parameters. These attacks can be client-side or server-side. Client-side attacks encompass data poisoning attack (DPA) and model poisoning attack (MPA) tactics. On the server side, the attacker can only implement MPA. DPA aims to manipulate the training data, while MPA directly alters model parameters. To implement the DPA attack, we apply the \textit{targeted label flipping (TLF)} method~\cite{nicolae2018adversarial}, which aims to make the model misclassify specific backdoored inputs and maintain the model performance on the other inputs. We also applied an \textit{untargeted label flipping (ULF)} attack that introduced random misclassifications. Regarding the MP, we implement \textit{client-side sign flipping attacks (CSF)} and \textit{server-side sign flipping attacks (SSF)} that flip the sign of the model parameters. Fig.~\ref{fig:AttackModel} shows the attack models during inference time and training time.



\subsection{Adversarial Defense on HFL Model}
Defenses against adversarial attacks can be broadly classified into two categories: \textit{data-driven} and \textit{model-driven} defenses~\cite{tian2022comprehensive}. Data-driven defenses involve detecting adversarial attacks in the data or enhancing the quality of the data corrupted by the attack to improve the performance of the model. These defense methods are typically agnostic to the learning architecture~\cite{tian2022comprehensive}. Model-driven defenses involve building models that are robust to adversarial attacks. 

In this work, to study the architectural impact of HFL on the efficacy of defense methods, we only implement model-driven defense methods that reconstruct the trained model to make it more robust. Thus, we implement \textit{Neural cleanse (NC)} ~\cite{nicolae2018adversarial}, a defense method that cleans the neural network from the neurons that are possibly affected by an attack. This method helps mitigate the impact of a TLF backdoor attack and produces a new, robust model. NC can be applied to the global model on the cloud server before it is sent to the clients. We also implement a well-known defense called \textit{adversarial training (AT)}~\cite{li2023federated}. AT is the process of retraining the model with adversarial examples to make the model recognize these examples and classify them correctly, even in the presence of perturbations. In the context of HFL, we can call it \textit{adversarial hierarchical federated training}. Each client implements local AT and collaborates with clients during adversarial hierarchical federated training to construct a robust global model against adversarial attacks in inference time.

\subsection{Experiment Design} \label{exp_des}
We conduct experiments to assess the impact of adversarial attacks on HFL models (3-level HFL and 4-level HFL) and compare the performance of HFL models under various attacks and defense mechanisms alongside CML and traditional FL approaches (2-level FL). Our code is available on GitHub\footnote{\url{https://github.com/dalqattan/SecHFL}}. The experimental settings are summarized as follows:

\textit{Dataset.} We use three popular image classification datasets: mnist, fashion-mnist, and cifar-10. Each dataset contains 60,000 images (of which 50,000 images are in the training set and 10,000 images are in the test set) categorized into 10 classes. To simulate non-IID real-world scenarios, the images of the training set are split according to the Dirichlet distribution. We use state-of-the-art implementation of attack and defense methods from~\cite{nicolae2018adversarial}.

\textit{HFL model.}
We consider a population of smart devices representing client nodes distributed across a city that implements image classification tasks. A group of 100 clients exists that engage in communication with the server for the purpose of image classification model training. We assume that the client selected for participation remains constant throughout the training process. Every client trains a local classifier model to classify the images. Regarding the server nodes, there is one cloud server in each learning paradigm at level 0. The cloud server performs the FedAvg aggregation rule for 20 aggregation rounds. We assume that the cloud server is highly secure and has never been compromised during the learning process. On the other hand, edge servers in HFL have different characteristics. In 3L-HFL, there are 20 regional edge servers that are distributed at the same level and connected directly with the cloud server and directly with 5 clients in their coverage area. Each regional edge server performs the FedAvg aggregation rule for two aggregation rounds. The 4L-HFL has similar settings to the 3L-HFL; however, there are 4 edge servers distributed at the same level between the cloud server and the regional edge servers. Each edge server communicates with five regional edge servers and performs the FedAvg aggregation rule for three aggregation rounds. The total aggregation round of regional edge servers is 40 and 120 rounds for 3L-HFL and 4L-HFL, respectively.

\textit{Client local training model.} We use two different convolutional neural network (CNN) architectures for the client's local classifier model for the three datasets. For mnist and fashion-mnist, we deploy a CNN with two 3x3 convolution layers (the first with 32 channels, the second with 64, each followed by 2x2 max-pooling), a fully connected layer with 512 units and ReLu activation, and a final softmax output layer with 10 outputs. For cifar10, A CNN with two 3x3 convolution layers with 32 channels followed by 2x2 max pooling, another two 3x3 convolution layers with 64 channels followed by 2x2 max pooling, a fully connected layer with 512 units and ReLu activation, and a final softmax output layer with 10 outputs. Each client employs categorical cross-entropy as their loss function and utilizes the optimizer that implements the Adam algorithm to update their local model depending on the loss function. For the mnist and fashion-mnist dataset, the batch size was set to 32 and the number of epochs was set to 1. For the cifar10 datasets, the batch size was set to 64, and the number of epochs was set to 6. 

\textit{Malicious Node.} If a client is compromised, the client could act maliciously by implementing DPA or MPA. We evaluate the performance of the model while the number of malicious clients is 1, 5, and 10. We also evaluate the models when all of the malicious clients are located in the overlapping area of two regional edge servers. We indicate the model that considers the overlapping area with the letter 'O' (3-level HFL-O and 4-level HFL-O). We also assume that regional edge servers can be compromised and act maliciously by implementing MPA, whereas other edge servers are highly secure. We evaluate the performance of the model while the number of malicious servers is 1, 5, and 10.

\textit{Evaluation Metrics.} We include the Misclassification Rate (MR) and the Targeted Attack Success Rate (TASR) to assess attack efficiency and defense effectiveness. The Misclassification Rate (MR) can be formulated as:
\begin{equation}
MP = \frac{1}{n}\sum_{i=1}^{n} \mathbb{I}( f(x_{i}^{'})\neq {y_{i}}),
\end{equation}
where $n$ is a number of image examples, $f(x_{i}^{'})$ is the aggregated model's output (global model output for HFL or centralized model output for CML) over input $x_i^{'}$ which is clean input ${x_i}$ for training-time attacks and adversarial input $x_{i}^{adv}$ for inference-time attacks, $y_i$ is ground truth, and $\mathbb{I}(\cdot,\cdot)$ is an indicator function that returns $1$ if model's output does match with the ground truth.

Similarly, TASR can be formulated as:
\begin{equation}
TASR = \frac{1}{n}\sum_{i=1}^{n} \mathbb{I}( f(x_{i}^{adv})={y_{i}^{adv}} ~|~ y_{i}^{adv}\neq y_{i}),
\end{equation}
where $f(x_{i}^{adv})$ is the aggregated model's output over adversarial input $x_i^a$ for a targeted adversarial attack label $y_i^{adv}$ in a backdoor attack, and $\mathbb{I}(\cdot,\cdot)$ is an indicator function that returns $1$ if model's output on attacked input matches with targeted adversarial attack label.

\section{Results and Discussion} \label{sec:results}
\subsection{Baseline performance: HFL model under no attacks}
\begin{figure}[t]
\centering
\includegraphics[width=12cm]{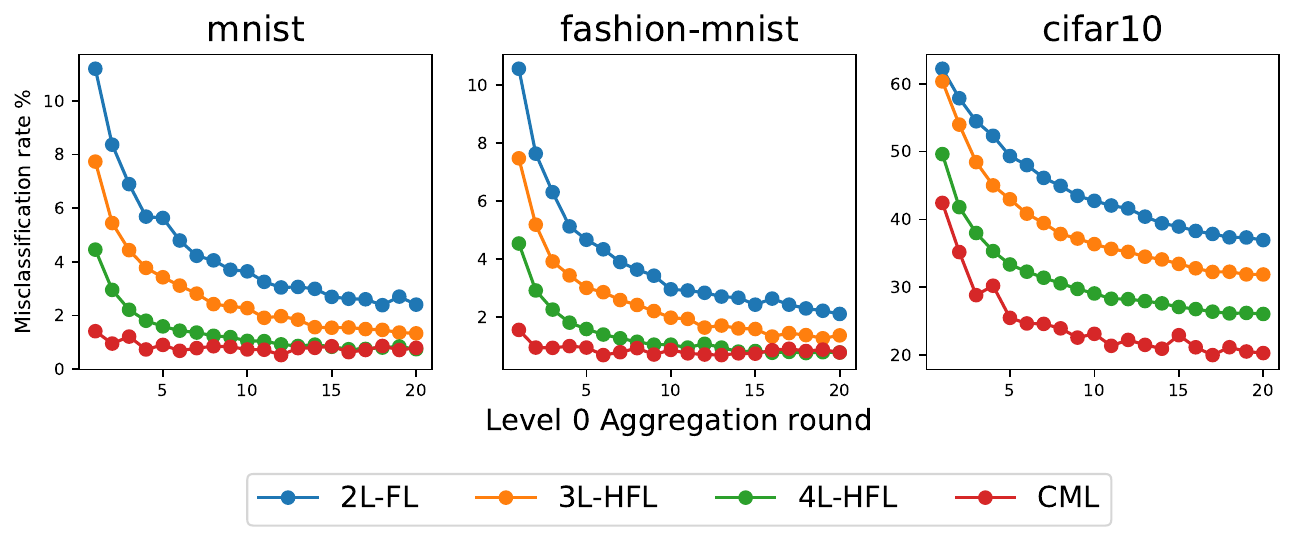}
\caption{Baseline performance: HFL models performance without adversarial attacks.}
\label{fig:models_performance}
\end{figure}

This section compares the performance of four models: a centralized machine learning model (CML), a 2-level FL, a 3-level HFL, and a 4-level HFL. As shown in Fig.~\ref{fig:models_performance}, the CML model maintains consistently high accuracy across 20 global aggregation rounds over each dataset. The 4-level HFL model demonstrates notably high performance, showcasing the potential advantages of hierarchical architecture in FL. The 3-level HFL model presents an intermediary performance between 4-level HFL and 2-level HFL models, showing how hierarchical architecture impacts FL. HFL architecture enhances model update efficiency and potentially leads to faster convergence. In contrast, the 2-level FL model shows inferior performance.

\subsection{Models performance under Inference-time attacks and defense}
\paragraph{Impact of the attacks.}
\begin{figure}[h]
\centering
\includegraphics[width=12cm]{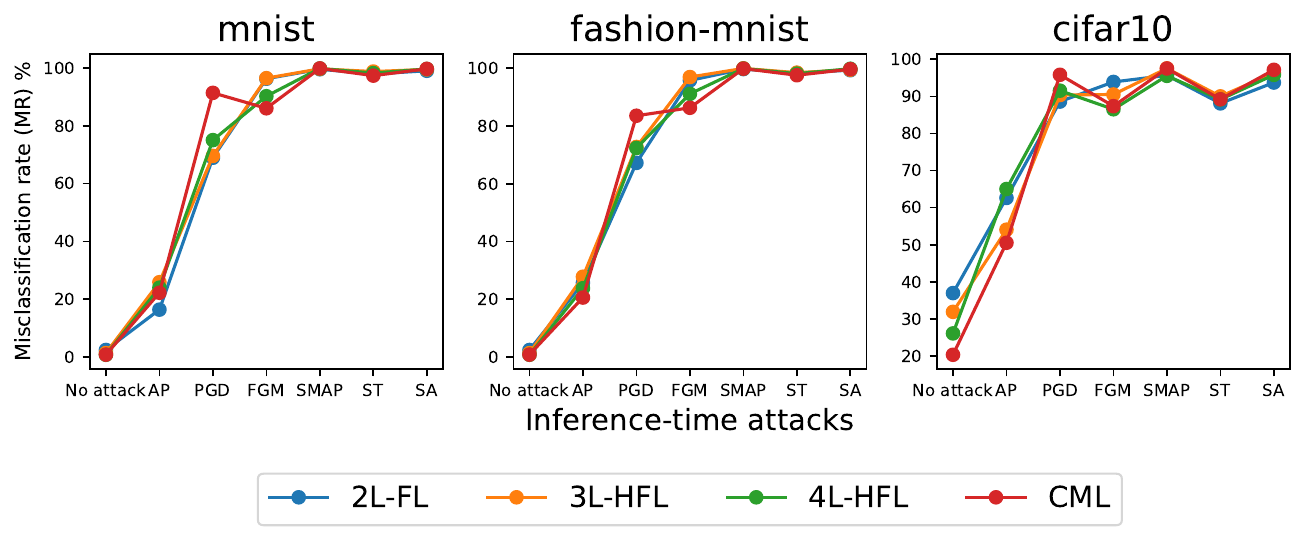}
\caption{Models performance under inference-time adversarial attacks.}
\label{fig:ITA_impact}
\end{figure}

We assessed the effectiveness of the models under attack by calculating MR. The outcomes are presented in Figure \ref{fig:ITA_impact}.
Upon analyzing the MR, it becomes evident that the MR of all models trained on the same dataset exhibits a high degree of similarity. However, the impact of each type of attack can vary. Adversarial patch attacks demonstrate the lowest impact. All the other attacks lead to a high MR ranging between 80\% to 100\%. As highlighted in \cite{kairouz2021advances}, in cases when the models, optimization methods, and the poisoned test dataset are identical, the effects of attacks on accuracy are likely to be comparable for both centralized machine learning and federated learning models. However, the reason for studying the impact of inference-time attacks in HFL is that many defenses against inference-time attacks are implemented during training. Thus, it is crucial to study the architectural impact on the model's robustness against inference-time attacks.

\paragraph{Adversarial training (AT) defense against inference-time attack.} \label{AT}
We adversarially trained all models using data generated by inference-time attacks to enhance their robustness. The effectiveness of these adversarially trained models was evaluated by measuring MR, as shown in Table~\ref{tab:AT}. In general, the MR dropped significantly across all models. While adversarially trained FL models demonstrate comparable MR to CML models, HFL models, especially the 4-level architecture, show even lower MR, suggesting higher resistance to attacks.
\begin{table}[h!]
\centering
\caption{Robustness of models (performance as per minimizing MR) due to AT (defense). The number in bold is the best defense among FL architectures}\label{tab:AT}
\begin{tabular}{llrrrrrrr}
\toprule
Dataset   & Model  & AP  & PGD~ & FGSM & JSMA     & ~~ST~~   & SA    & Average   \\ \midrule
\multirow{4}{*}{mnist}    & 2L FL  & 8.82      & 2.27      & 2.97      & 2.05      & 2.66 & 4.93      & 3.95      \\ 
     & 3L HFL & 15.64     & 1.48      & 2.13      & 1.89      & 11.87      & 7.34      & 6.73      \\  
     & 4L HFL & \textbf{5.35}  & \textbf{0.92} & \textbf{1.6}  & \textbf{0.97}  & \textbf{1.15}  & \textbf{1.8}  & \textbf{1.96} \\  
     & CML    & 5.16     & 0.96      & 1.71      & 0.84     & 6.73 & 2.13      & 2.92      \\ \midrule
\multirow{4}{*}{fashion-mnist} & 2L FL  & 12.58     & 2.31      & 2.88      & 2.56      & 2.22 & 4.98      & 4.59      \\  
     & 3L HFL & 8.29      & 1.32      & 1.74      & 1.59      & \textbf{1.68}  & 3.6 & 3.04      \\  
     & 4L HFL & \textbf{5.65} & \textbf{0.9}  & \textbf{1.27} & \textbf{1.07}  & 6.58 & \textbf{2.13} & \textbf{2.93} \\  
     & CML    & 5.86      & 1.06      & 1.91      & 1.01     & 9.52 & 2.75      & 3.69      \\ \midrule
\multirow{4}{*}{cifar10}  & 2L FL  & 44.28     & 49.67     & 43.1      & 41.14     & \textbf{60.42} & 51.92     & 48.42     \\  
     & 3L HFL & 39.95     & 47.51     & 39.05     & 37.53     & 56.17      & 43.2      & 43.90     \\ 
     & 4L HFL & \textbf{34.89} & \textbf{41.79} & \textbf{38.15} & \textbf{32.16} & 62.75      & \textbf{39.81} & \textbf{41.59} \\  
     & CML    & 28.56    & 27.35    & 30.17    & 22.36    & 60.94      & 29.23    & 33.10    \\ 
\bottomrule
\end{tabular}
\end{table} 

\begin{figure}[!t]
\centering
\includegraphics[width=12cm]{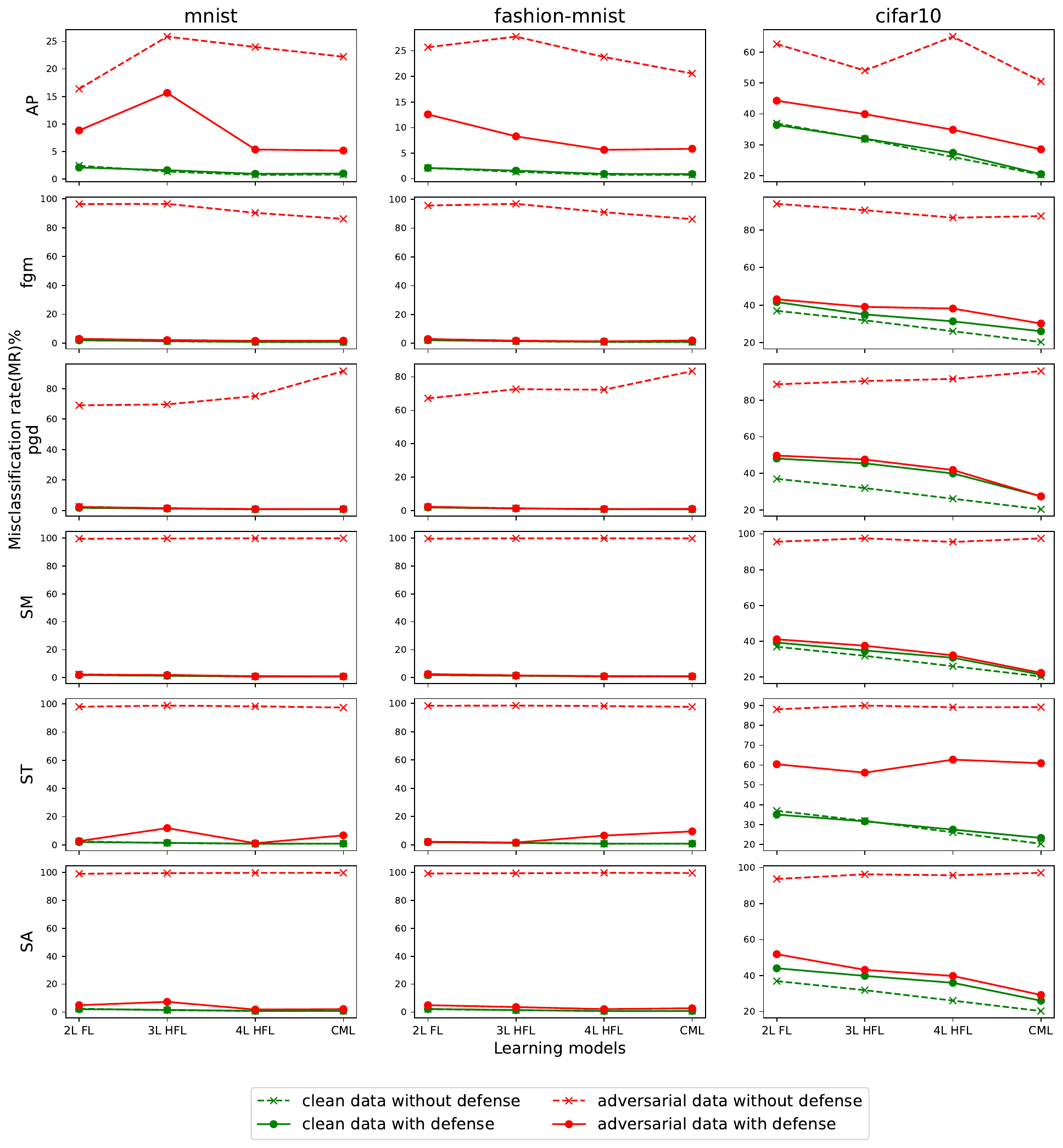}
\caption{Model's performance under Inference-time attacks and adversarial Training defense}
\label{fig:ITA-AT}
\end{figure}

Fig.~\ref{fig:ITA-AT} shows the improved MR of robust models (red solid line) achieved through adversarial training compared to vulnerable models (red dashed line). However, a drawback of direct adversarial training adoption is observed with increased dataset complexity (cifar10), leading to higher MR for clean data, emphasizing the need for further research on complex, large-scale datasets.

\subsection{Models performance under Training-time attacks and defense}\label{TTA} 
\begin{figure}[!t]
\centering
\includegraphics[width=12cm]{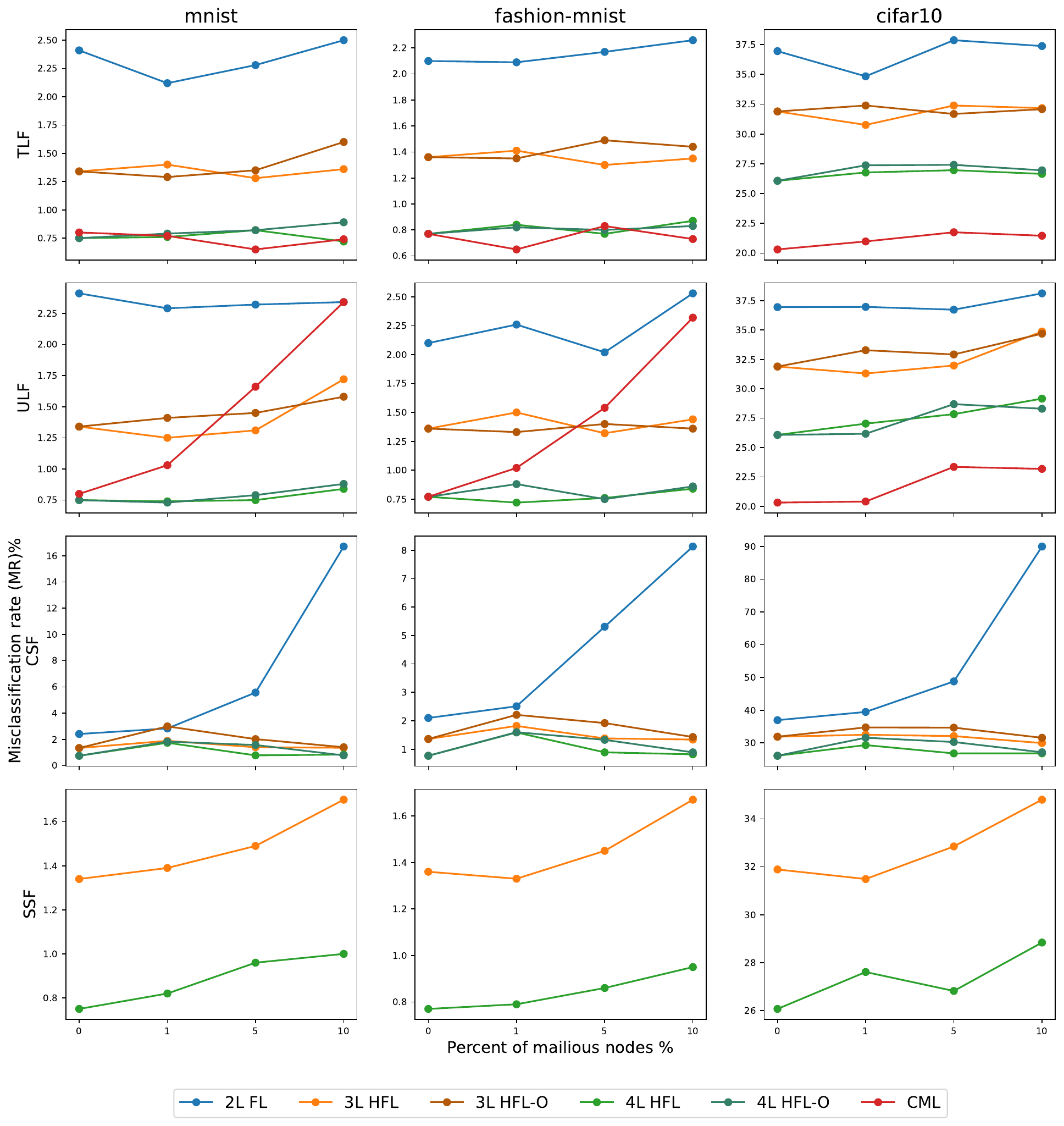}
\caption{Model's performance under Training-time attacks}
\label{fig:TTA}
\end{figure}

Fig.~\ref{fig:TTA} shows the consequences of training-time attacks on five distinct FL models that possess varied degrees of hierarchy and compromised nodes across different clean test datasets. The x-axis shows the number of compromised nodes (0, 1, 5, and 10), while the y-axis signifies the impact of the attack, reflecting the increase in MR resulting from the training-time attacks. The letter 'O' in the model name indicates that all the malicious clients are located in an overlapping area of two regional servers.

\subsubsection{The impact of client-side attacks (data poisoning)}
We study both targeted and untargeted attacks on HFL as follows:
\paragraph{Targeted label flipping (TLF) with backdoor attack.}
The targeted backdoor attack has two aims. First, to maintain the model's performance on clean data. Second, to make the model misclassify the targeted label as a desired label.

TLF backdoor attack result in Fig.~\ref{fig:TTA} shows that the MR for all three clean test datasets remains relatively stable across different percentages of malicious clients. This stability suggests that the presence of malicious clients has little impact on the model’s performance, even when malicious clients are located in the overlapping areas of two servers. This indicates that the attacker fully achieved the first aim of not influencing the model’s performance on clean test datasets.

\begin{figure}[t]
\centering
\includegraphics[width=12cm]{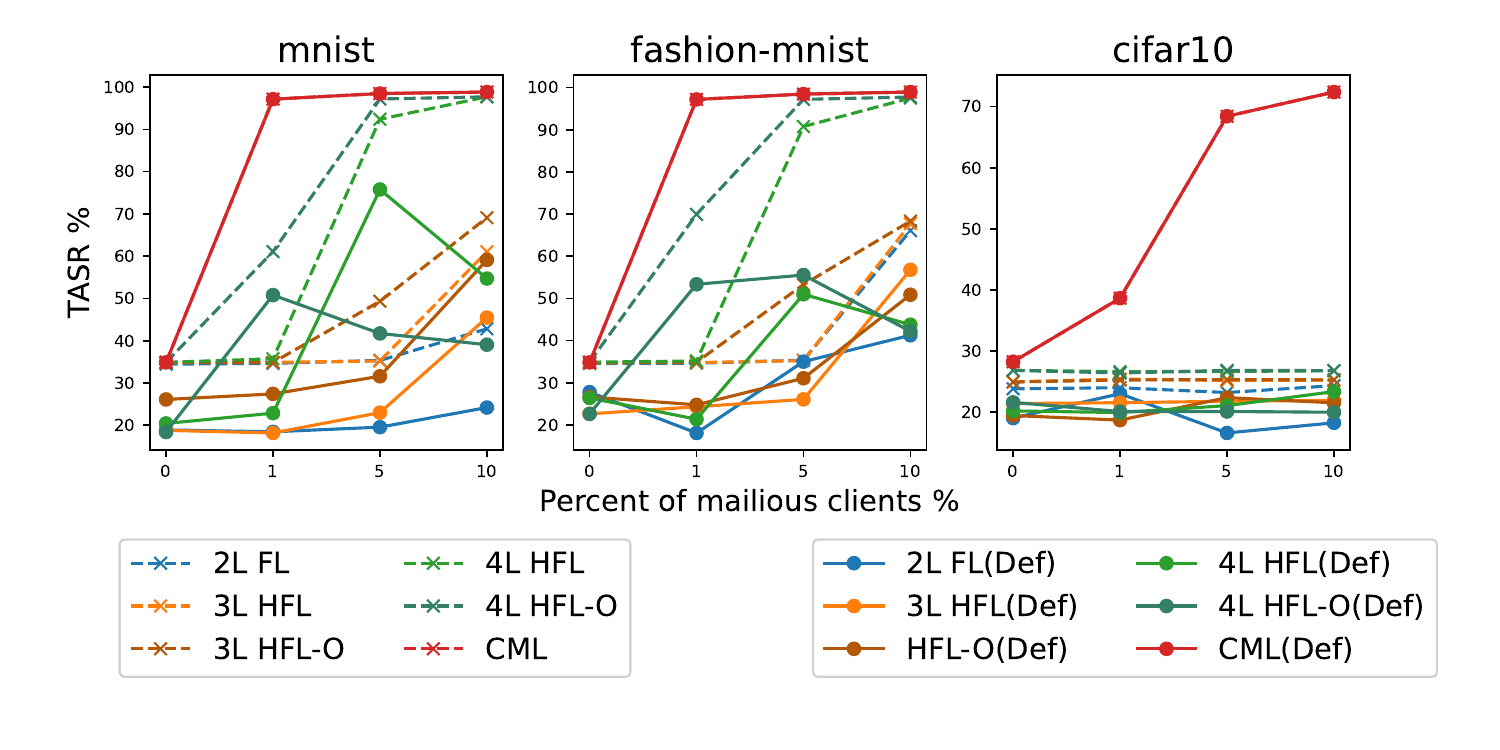}
\caption{Success rate of backdoor attacks before (dashed line) and after (solid line) neural cleanser defence.}
\label{fig:TTA-BD-with-defence}
\end{figure}

The analysis of the second aim is shown in Fig.~\ref{fig:TTA-BD-with-defence}. From Fig.~\ref{fig:TTA-BD-with-defence}, we observe that TASR increases with the percentage of malicious clients for all models. CML model shows a notably high TASR, indicating vulnerability to backdoor attacks. Among FL models, the 4-level model consistently demonstrates the highest vulnerability to backdoor attacks, followed by the 3-level model and then the 2-level model. This suggests that increased complexity in FL models does not necessarily correlate with improved security against backdoor attacks. Malicious clients located in the overlapping area further amplify the potency of backdoor attacks, underscoring the importance of tailored security measures required in FL environments.

The neural cleanse (NC) method offers a robust defense against backdoor attacks in FL models, significantly reducing TASR and enhancing overall model security and robustness. Fig.~\ref{fig:TTA-BD-with-defence} (solid lines) shows the effectiveness of this method across various FL models, showcasing a substantial reduction in TASR compared to scenarios without defense mechanisms (dashed line).

Despite these improvements, CML models still show higher TASR values, highlighting their inherent vulnerabilities to backdoor attacks compared to FL models. This underscores the inherent vulnerabilities of CML systems to backdoor attacks and emphasizes the relative resilience of FL models when equipped with NC defense mechanisms. The degree of improvement in TASR varies based on factors like dataset complexity, percentage of malicious clients, and model architecture, emphasizing the need for adaptive defense strategies tailored to specific attack scenarios.

\paragraph{Untargeted random label flipping (ULF) attack.}
As shown in Fig.~\ref{fig:TTA}, CML models suffer amplified effects from such attacks as increased compromised clients. However, FL and HFL are less impacted. For instance, in the mnist dataset, with 10 compromised clients, the MR increases by only 0.2\% compared to models without attacks. Although HFL has slightly higher susceptibility due to server coverage, its impact remains minimal. FL's resilience is attributed to its client selection mechanism, where only a small proportion of clients are chosen per round, reducing the likelihood of selecting compromised clients.  Moreover, to reduce FL and HFL accuracy, more than 10 clients must be compromised, necessitating a high-budget attack. Furthermore, imposing constraints on local dataset sizes effectively mitigates the occurrence of poisoned data, offering an efficient defense against untargeted attacks. This observation is consistent with findings presented in~\cite{shejwalkar2022back}, further supporting the resilience of FL in real-world scenarios.

\subsubsection{Impact of client-side attacks (model poisoning)}
In model poisoning [Client-side Sign flipping (CSF)], we only evaluate the result for FL models. This is because model poisoning is not commonly applied in CML. Regarding model poisoning attacks, Fig.~\ref{fig:TTA} shows that all five FL models show minimal increases in MR, indicating resilience against such attacks. However, the 2-level FL model displays significant vulnerability when 10 clients are compromised, as observed in~\cite{shejwalkar2022back}. Conversely, the 3-level and 4-level HFL models show stronger performance, attributed to their hierarchical aggregation process, which mitigates the impact of individual clients. Even when all compromised clients strategically overlap two servers, HFL models show lesser MR impact compared to the 2-level model. These findings underscore the importance of hierarchical structure in mitigating model poisoning effects, suggesting the need for enhanced security measures for the 2-level FL model.

\subsubsection{The impact of server-side attacks(model poisoning)}
In comparing server-side sign-flipping (SSF) attacks between 3-level and 4-level HFL models, we observe in Fig.~\ref{fig:TTA} that the 4-level model consistently shows lower MR across all datasets, indicating greater resilience to model poisoning. The impact increases with the number of compromised servers yet remains negligible, with both models showing only a slight increase in MR even when 10 servers are compromised. Specifically, the MR increase for the 3-level model does not exceed 0.4\% for mnist and fashion-mnist datasets, while for CIFAR-10, both models show only a 3\%-4\% increase in MR. These results highlight the robustness of HFL models against server-side attacks, particularly for the 4-level architecture.

From the results of a systematic analysis of HFL security, we observe that, in the context of ITAs, HFL models show varying degrees of susceptibility to adversarial perturbations during the inference phase. These findings underscore the importance of evaluating model robustness against a diverse range of ITAs to ensure reliable performance in real-world scenarios. AT emerges as a promising defense strategy, effectively enhancing model robustness against such attacks. Notably, adversarially trained FL models, especially those HFL models, demonstrate competitive misclassification rates compared to CML. The 4-level HFL architecture, in particular, shows notable resilience in adversarial training, suggesting its efficacy in mitigating adversarial attacks.

Regarding TTAs, the 4-level HFL model shows the highest vulnerability to TLF attack, particularly when malicious clients are positioned in overlapping areas of regional servers. However, our investigation also assesses the effectiveness of defense mechanisms, such as the NC method, in mitigating TLF attacks within HFL systems. The NC method significantly reduces the TASR, enhancing the overall security posture of HFL models.

Moreover, FL and HFL models show greater resilience to ULF attacks, with minimal MR increases even when a considerable number of clients are compromised. This resilience can be attributed to the multi-level aggregation inherent in HFL, which effectively smooths out the impact of outliers introduced by such attacks. This ability to recover from attacks further underscores the robustness of HFL in real-world deployment scenarios.

\section{Conclusion} \label{sec:con}
Our investigation reveals that hierarchical federated learning (HFL) is resilient to untargeted data poisoning due to its hierarchical structure. However, targeted attacks, like backdoors, exploit architectural nuances, particularly when malicious clients strategically position themselves in the overlapping coverage area of regional edge servers. This highlights the need for further research in HFL security. Nonetheless, HFL shows promise in enhancing adversarial training to counter inference-time attacks. Future efforts should focus on developing tailored defense mechanisms to mitigate risks, bolstering the overall security and reliability of HFL systems for broader applications.

\section*{Acknowledgements}
This research was supported by the Technical and Vocational Training Corporation (TVTC) through the Saudi Arabian Culture Bureau (SACB) in the United Kingdom and the EPSRC-funded project National Edge AI Hub for Real Data: Edge Intelligence for Cyber-disturbances and Data Quality (EP/Y028813/1).


\bibliographystyle{splncs04}
\bibliography{icannref}

\end{document}